\documentclass[11pt,a4paper]{article}
\usepackage[hyperref]{emnlp2018}
\usepackage{times}
\usepackage{latexsym}
\usepackage{booktabs}
\usepackage{url}
\usepackage{graphicx}
\usepackage[utf8]{inputenc}
\usepackage{enumitem}
\usepackage{microtype}

\newlist{examples}{enumerate}{2}
\setlist[examples]{noitemsep,resume}
\setlist[examples,1]{label=(\arabic*)}
\setlist[examples,2]{label=\alph*.}

\setlist[itemize]{noitemsep}

\usepackage{pifont}
\newcommand{\cmark}{\ding{51}}%
\newcommand{\xmark}{\ding{55}}%

\aclfinalcopy 

\title{Automatic Reference-Based Evaluation of Pronoun Translation\\Misses the Point}

\author{Liane Guillou \\
  University of Edinburgh \\
  School of Informatics \\
  Scotland, United Kingdom \\
  {lguillou@inf.ed.ac.uk} \\\And
  Christian Hardmeier \\
  Uppsala University \\
  Dept.\ of Linguistics and Philology \\
  Uppsala, Sweden \\
  {christian.hardmeier@lingfil.uu.se} \\}

\date{}

\begin{document}
\maketitle

\begin{abstract}
We compare the performance of the APT and AutoPRF metrics for pronoun translation against 
a manually annotated dataset comprising human judgements as to the correctness of translations 
of the PROTEST test suite. 
Although there is some correlation with the human judgements, a range 
of issues limit the performance of the automated metrics. 
Instead, we recommend the use of semi-automatic metrics and test suites in place of fully automatic metrics.
\end{abstract}

\section{Introduction}

As the general quality of machine translation (MT) increases, there is a growing
interest in improving the translation of specific linguistic phenomena. A case
in point that has been studied in the context of both statistical
\cite{HardmeierThesis,GuillouThesis,LoaicigaThesis} and neural MT
\cite{Bawden2017} is that of pronominal anaphora. In the simplest case,
translating anaphoric pronouns requires the generation of corresponding word
forms respecting the grammatical constraints on agreement in the target
language, as in the following English-French example, where the correct form of
the pronoun in the second sentence varies depending on which of the (equally
correct) translations of the word \textit{bicycle} was used in the first:
\begin{examples}
\item
\begin{examples}
  \item I have a bicycle. \textbf{It} is red.
  \item J'ai un v\'{e}lo. \textbf{Il} est rouge. [ref]
  \item J'ai une bicyclette. \textbf{Elle} est rouge. [MT]
\end{examples}
\end{examples}
However, the problem is more complex in practice because there is often no $1:1$
correspondence between pronouns in two languages. This is easily demonstrated at
the corpus level by observing that the number of pronouns varies significantly
across languages in parallel texts \cite{Mitkov:2003}, but it tends to be
difficult to predict in individual cases.

In general MT research, significant progress was enabled by the invention of
automatic evaluation metrics based on reference translations, such as BLEU
\cite{Papineni:2002}. Attempting to create a similar framework for efficient
research, researchers have proposed automatic reference-based evaluation
metrics specifically targeting pronoun translation
\cite{Hardmeier2010,Werlen2017}. In this paper, we study the performance of
these metrics on a dataset of English-French translations and investigate to
what extent automatic evaluation based on reference translations can provide
useful information about the capacity of an MT system to handle pronouns. Our
analysis clarifies the conceptual differences between AutoPRF and APT,
uncovering weaknesses in both metrics, and investigates the effects of the
alignment correction heuristics used in APT. By using the fine-grained PROTEST
categories of pronoun function, we find that the accuracy of the automatic
metrics varies across pronouns of different functions, suggesting that certain
linguistic patterns are captured better in the automatic evaluation than others.
We argue that fully automatic wide-coverage evaluation of this phenomenon is
unlikely to drive research forward, as it misses essential parts of the problem
despite achieving some correlation with human judgements. Instead,
semi-automatic evaluation involving automatic identification of correct
translations with high precision and low recall appears to be a more achievable
goal. Another more realistic option is a test suite evaluation with a very
limited scope.

\section{Pronoun Evaluation Metrics for MT}


Two reference-based automatic metrics of pronoun translation have been proposed
in the literature. The first \cite{Hardmeier2010} is a variant of precision,
recall and F-score that measures the overlap of pronouns in the MT output with
a reference translation. It lacks an official name, so we refer to it as
\emph{AutoPRF} following the terminology of the DiscoMT 2015 shared task
\cite{Hardmeier2015b}. The scoring process relies on a word alignment between
the source and the MT output, and between the source and the reference
translation. For each input pronoun, it computes a \emph{clipped count}
\cite{Papineni:2002} of the overlap between the aligned tokens in the reference
and the MT output. The final metric is then calculated as the precision, recall
and F-score based on these clipped counts.

\newcite{Werlen2017} propose a metric called \textit{Accuracy of Pronoun
Translation (APT)} that introduces several innovations over the previous work.
It is a variant of accuracy, so it counts, for each source pronoun, whether its
translation can be considered correct, without considering multiple alignments.
Since word alignment is problematic for pronouns, the authors propose an
heuristic procedure to improve alignment quality. Finally, it introduces the
notion of pronoun equivalence, assigning partial credit to pronoun translations
that differ from the reference translation in specific ways deemed to be
acceptable. In particular, it considers six possible cases when comparing the
translation of a pronoun in MT output and the reference. The pronouns may
be: (1) \textit{identical}, (2) \textit{equivalent}, (3) \textit{different/incompatible},
or there may be no translation in: (4) the MT output, (5) the reference, (6)
either the MT output or the reference. Each of these cases may be assigned a
weight between 0 and 1 to determine the level of \textit{correctness}.

\section{The PROTEST Dataset}
We study the behaviour of the two automatic metrics using the PROTEST test suite
\cite{Guillou2016}. It comprises 250 hand-selected \textit{personal} pronoun
tokens taken from the \textit{DiscoMT2015.test} dataset
\cite{DiscoMT2015TestSet} and annotated according to the ParCor guidelines
\cite{ParCor2014}. Pronouns are first categorised according to their
\textit{function}:

\hangindent=\parindent
\textit{anaphoric}: I have a bicycle. \textbf{It} is red.\\
\textit{event}: He lost his job. \textbf{It} was a shock.\\
\textit{pleonastic}: \textbf{It} is raining.\\
\textit{addressee reference}: \textbf{You}'re welcome.

and then subcategorised according to morphosyntactic criteria, whether the
antecedent is a group noun, whether the ancedent is in the same or
a different sentence, and whether an addressee reference pronoun 
refers to one or more specific people (deictic) or to people in general
(generic).

Our dataset contains human judgements on the performance of eight MT systems on
the translation of the 250 pronouns in the PROTEST test suite. The systems
include five submissions to the DiscoMT~2015 shared task on pronoun translation
\cite{Hardmeier2015b} -- four phrase-based SMT systems \textsc{auto-postEDIt}
\cite{Guillou2015}, \textsc{uu-hardmeier} \cite{Hardmeier2015b}, \textsc{idiap}
\cite{Luong2015}, \textsc{uu-tiedemann} \cite{Tiedemann2015}, a rule-based
system \textsc{its2} \cite{Loaiciga2015}, and the shared task baseline (also
phrase-based SMT). Two NMT systems are included for comparison: \textsc{limsi}
\cite{Bawden2017} and \textsc{nyu} \cite{NYUNMT}. 

Manual evaluation was conducted using the PROTEST graphical user interface 
and accompanying guidelines \cite{Hardmeier2016}. The annotators were asked to
make judgements (correct/incorrect) on the translations of the pronouns and
antecedent heads whilst ignoring the correctness of other words (except in cases
where it impacted the annotator's ability to make a judgement). The annotations
were carried out by two bilingual English-French speakers, both of whom are
native speakers of French. Note that our human judgements differ in important
ways from the human evaluation conducted for the same set of systems at
DiscoMT~2015 \cite{Hardmeier2015b}, which was carried out by non-native speakers
over an unbalanced data sample using a gap-filling methodology.

\section{Accuracy versus Precision/Recall}

There are three ways in which APT differs from AutoPRF: the scoring statistic,
the alignment heuristic in APT and the definition of pronoun equivalence.

APT is a measure of accuracy: It reflects the proportion of source pronouns for
which an acceptable translation was produced in the target. AutoPRF, by
contrast, is a precision/recall metric on the basis of clipped counts. 
The reason for using precision and recall given by
\newcite{Hardmeier2010} is that word alignments are not $1:1$, and each pronoun
can therefore be linked to multiple elements in the target language, both in the
reference translation and in the MT output. Their metric is designed to account
for all linked words in such cases.

To test the validity of this argument, we examined the subset of examples in our
English--French dataset giving rise to a clipped count greater than 1 and found
that these examples follow very specific patterns. All 143 cases included
exactly one personal pronoun. In 99 cases, the additional matched word was the
complementiser \textit{que} `that'. In 31 and 4 cases, respectively, it was a
form of the auxiliary verbs \textit{avoir} `to have' and \textit{être} `to be'.
One example matched both \emph{que} and a form of \textit{être}. Two had
reflexive pronouns, and one an imperative verb form. With the possible exception
of the two reflexive pronouns, none of this seems to be relevant to pronoun
correctness. We conclude that it is more reasonable to restrict the counts to a
single pronominal item per example. With this additional restriction, however,
the recall score of AutoPRF becomes equivalent to a version of APT without
equivalent pronouns and alignment correction. We therefore limit the remainder of
our study to APT.

\begin{table}[t]\centering\small
\begin{tabular}{lccccc}
\toprule
Score             & \multicolumn{2}{c}{\textbf{APT-A}} & \multicolumn{2}{c}{\textbf{APT-B}} & \textbf{PRO-}\\
Alig. corr.       & +                                  & --                                 & +               & --             & \textbf{TEST} \\
\midrule
Reference         & 1.000                              & 1.000                              & 1.000           & 1.000          & 0.920 \\
\textsc{baseline} & \textbf{0.544}                     & \textbf{0.536}                     & \textbf{0.574}  & \textbf{0.566} & 0.660 \\
\textsc{idiap}    & 0.496                              & 0.496                              & 0.528           & 0.528          & 0.660 \\
\textsc{uu-tied.} & 0.532                              & 0.532                              & 0.562           & 0.562          & \textbf{0.680} \\
\textsc{uu-hard.} & 0.528                              & 0.520                              & 0.556           & 0.548          & 0.636 \\
\textsc{postEDIt} & 0.492                              & 0.492                              & 0.532           & 0.532          & 0.668 \\
\textsc{its2}     & 0.436                              & 0.428                              & 0.462           & 0.454          & 0.472 \\
\textsc{limsi}    & 0.364                              & 0.364                              & 0.388           & 0.388          & 0.576 \\
\textsc{nyu}      & 0.424                              & 0.420                              & 0.456           & 0.452          & 0.616 \\
\bottomrule
\end{tabular}
\caption{Comparison of APT scores with human judgements over the PROTEST test suite}
\label{table:APTandPROTEST}
\end{table}

\section{Effects of Word Alignment}

APT includes an heuristic alignment correction procedure to mitigate errors in
the word alignment between a source-language text and its translation (reference
or MT output). We ran experiments to assess the correlation of 
APT with human judgements, with and without the alignment heuristics.


Table \ref{table:APTandPROTEST} displays the APT results, with and without the
alignment heuristics, and the proportion of pronouns in the PROTEST test suite
marked as correctly translated. We computed APT scores for two different weight
settings:\footnote{Personal recommendation by Lesly Miculicich
Werlen.} \textbf{APT-A} uses weight 1 for \textit{identical} matches and 0
for all other cases. \textbf{APT-B} uses weight 1 for \textit{identical}
matches, 0.5 for \textit{equivalent} matches and 0 otherwise.

There is little difference in the APT scores when we consider the use of
alignment heuristics. This is due to the small number of pronouns for which
alignment improvements are applied for most systems (typically 0--9 per system).
The exception is the \textsc{its2} system output for which 18 alignment
improvements are made. For the following systems we observe  a very small
increase in APT score for each of the two weight settings we consider, when
alignment heuristics are applied: \textsc{uu-hardmeier} (+0.8), \textsc{its2}
(+0.8), the \textsc{baseline} (+0.8) and \textsc{nyu} (+0.4). However, these
small improvements are not sufficient to affect the system rankings.

\begin{table}[t]\centering\small
\setlength\tabcolsep{0.2em}
\begin{tabular}{lccrr}
\toprule
& c1 & c2 & Pearson & Spearman\\
\midrule
With alignment & 1 & 0 & 0.907 & 0.778 \\
heuristics & 1 & 0.5 & 0.913 & 0.803 \\
\midrule
Without alignment & 1 & 0 & 0.913 & 0.778 \\
heuristics & 1 & 0.5 & 0.919 & 0.803 \\
\bottomrule
\end{tabular}
\caption{Correlation of APT and human judgements}
\label{table:CorrelationScores}
\end{table}



\section{Metric Accuracy per Category}

\begin{table}\centering\small
\setlength\tabcolsep{0.2em}
\begin{tabular}{lrrrrrr@{\,/\,}rr}
\toprule
& \multicolumn{3}{c}{\textbf{APT}} & \multicolumn{2}{c}{\textbf{Human}} & \multicolumn{3}{c}{\textbf{Disagreement}} \\
\textbf{Category} & \multicolumn{3}{c}{\textbf{Cases}} & \multicolumn{2}{c}{\textbf{Assess.}} & \multicolumn{3}{c}{} \\
                              & \multicolumn{1}{c}{1}     & \multicolumn{1}{c}{2}   & \multicolumn{1}{c}{3}   & \multicolumn{1}{c}{\cmark} & \multicolumn{1}{c}{\xmark} &    \multicolumn{2}{c}{} & \multicolumn{1}{c}{\%} \\
\midrule
Anaphoric                & \multicolumn{8}{c}{}\\
\quad intra sbj \textit{it}  & 112   & 13  & 68  & 133   & 60  & 42  & 193   & 21.8 \\
\quad intra nsbj \textit{it} & 52    & 1   & 25  & 65    & 13  & 12  & 78    & 15.4 \\
\quad inter sbj \textit{it}  & 99    & 17  & 95  & 130   & 81  & 56  & 211   & 26.5 \\
\quad inter nsbj \textit{it} & 18    & 0   & 7   & 6     & 19  & 12  & 25    & 48.0 \\
\quad intra \textit{they}    & 115   & 0   & 86  & 133   & 68  & 30  & 201   & 14.9 \\
\quad inter \textit{they}    & 117   & 0   & 94  & 118   & 93  & 43  & 211   & 20.4 \\
\quad sg \textit{they}       & 52    & 0   & 58  & 72    & 38  & 48  & 110   & 43.6 \\
\quad group \textit{it/they} & 45    & 0   & 35  & 57    & 23  & 26  & 80    & 32.5 \\
Event \textit{it}            & 125   & 38  & 89  & 157   & 95  & 56  & 252   & 22.2 \\
Pleonastic \textit{it}       & 155   & 49  & 46  & 216   & 34  & 40  & 250   & 16.0 \\
Generic \textit{you}         & 105   & 0   & 62  & 166   & 1   & 61  & 167   & 36.5 \\
Deictic sg \textit{you}      & 85    & 0   & 43  & 126   & 2   & 41  & 128   & 32.0 \\
Deictic pl \textit{you}      & 81    & 0   & 7   & 87    & 1   & 6   & 88    & 6.9 \\
\midrule
Total                        & 1,161 & 118 & 715 & 1,466 & 528 & 473 & 1,994 & 23.7 \\
\bottomrule
\end{tabular}
\caption{Number of pronouns marked as correct/incorrect in the PROTEST human judgements, as identical (1), equivalent (2), and incompatible (3) by APT, and the percentage of disagreements, per category}
\label{table:APTpercategory}
\end{table}

Like \citet{Werlen2017}, we use Pearson's and Spearman's correlation
coefficients to assess the correlation between APT and our human judgements
(Table \ref{table:CorrelationScores}). Although APT does correlate with the
human judgements over the PROTEST test suite, the correlation is weaker than
that with the DiscoMT gap-filling evaluations reported in \citet{Werlen2017}.
Table~\ref{table:APTandPROTEST} also shows that the rankings induced from
the PROTEST and APT scores are rather different.


We also study how the results of APT
(with alignment correction) interact with the categories in PROTEST. We consider a pronoun to be measured as \textit{correct} by APT if it is assigned 
case 1 (identical) or 2 (equivalent). Likewise, a pronoun 
is considered \textit{incorrect} if it is assigned case 3 (incompatible).
We compare the number of pronouns marked as correct/incorrect by APT and by the human judges, 
ignoring APT cases in which no judgement can be made: no 
translation of the pronoun in the MT output, reference or both, and pronouns 
for which the human judges were unable to make a judgement due to 
factors such as poor overall MT quality, incorrect word alignments, etc. 
The results of this comparison are displayed in Table \ref{table:APTpercategory}.

At first glance, we can see that APT disagrees with the human judgements for
almost a quarter (23.72\%) of the assessed translations. The distribution of the
disagreements over APT is very skewed and ranges from 9\% for case 1 to 34\% for
case 2 and 46\% for case 3. In other words, APT identifies correct pronoun
translations with good precision, but relatively low recall. We can also see
that APT rarely marks pronouns as equivalent (case 2).



APT performs particularly poorly on the assessment of pronouns belonging to the 
\textit{anaphoric inter-sentential non-subject ``it''} and 
\textit{anaphoric singular ``they''} categories. In general, there are three main problems affecting 
anaphoric pronouns (Table \ref{table:CommonErrors}). 
1) APT does not consider pronoun-antecedent head agreement so many \textit{valid alternative 
translations involving personal pronouns} are marked as incompatible (case 3), but as correct by the 
human judges.
2) Substitutions between pronouns are governed by much more complex rules than
the simple pronoun equivalence mechanism in APT suggests.
3) APT does not consider the use of \textit{impersonal pronouns}
such as \textit{c'} in place of the feminine personal pronoun \textit{elle} or
the plural forms \textit{ils} and \textit{elles}. 

\begin{table}[h]\centering\small
\setlength\tabcolsep{0.5em}
\begin{tabular}{lrrrr}
\toprule
\textbf{Category} & \textbf{V} & \textbf{E} & \textbf{I} & \textbf{O}\\
\midrule
Anaphoric & & & & \\
\hspace{3mm}intra--sent. subj. it & 26 & 9 & 7 & -- \\
\hspace{3mm}intra--sent. non--subj. it & -- & -- & -- & 12 \\ 
\hspace{3mm}inter--sent. subj. it & 32 & 5 & 19 & -- \\
\hspace{3mm}inter--sent. non--subj. it & 12 & -- & -- & -- \\
\hspace{3mm}intra--sent. they & 26 & -- & 2 & 2 \\
\hspace{3mm}inter--sent. they & 41 & -- & 2 & -- \\
\hspace{3mm}singular they & 47 & -- & -- & 1 \\
\hspace{3mm}group it/they & 24 & -- & -- & 2 \\
\midrule
Event it & -- & 16 & -- & 40 \\
\midrule
Pleonastic it & -- & 11 & -- & 29 \\
\bottomrule
\end{tabular}\\[1mm]
\textbf{V}: Valid alternative translation\qquad\textbf{I}: Impersonal translation\\
\textbf{E}: Incorrect equivalence\qquad \textbf{O}: Other
\caption{Common cases of disagreement for anaphoric, pleonastic, and event reference pronouns}
\label{table:CommonErrors}
\end{table}

As with anaphoric pronouns, APT incorrectly marks some \textit{pleonastic} and
\textit{event} translations as equivalent in disagreement with the human judges.
Other common errors arise from 1) the use of alternative translations marked as
incompatible by APT but correct by the human judges, for example \textit{il}
(personal) in the MT output when the reference contained the impersonal pronoun
\textit{cela} or \textit{\c{c}a} (25 cases for pleonastic, 6 for event), or 2)
the presence of \textit{il} in both the MT output and reference which APT marked
as identical but the human judges marked as incorrect (3 cases for pleonastic,
16 event).


Some of these issues could be addressed by incorporating knowledge of pronoun
function in the source language, pronoun antecedents, and the wider context of
the translation surrounding the pronoun. However, whilst we might be able to
derive language-specific rules for some scenarios, it would be difficult to come
up with more general or language-independent rules. For example, \textit{il} and
\textit{ce} can be anaphoric or pleonastic pronouns, but \textit{il} has a more
\textit{referential} character. Therefore in certain constructions that are
strongly pleonastic (e.g.\ clefts) only \textit{ce} is acceptable. This rule
would be specific to French, and would not cover other scenarios for the
translation of pleonastic \textit{it}. Other issues include the use of pronouns
in impersonal constructions such as \textit{il faut} [one must/it takes] in
which evaluation of the pronoun requires consideration of the whole expression,
or transformations between active and passive voice, where the perspective of
the pronouns changes.

\section{Conclusions}
Our analyses reveal that despite some correlation between APT and the human
judgements, fully automatic wide-coverage evaluation of pronoun translation
misses essential parts of the problem. Comparison with human judgements shows
that APT identifies good translations with relatively high precision, but fails
to reward important patterns that pronoun-specific systems must strive to
generate. Instead of relying on fully automatic evaluation, our recommendation is to emphasise high precision in the automatic metrics and implement
semi-automatic evaluation procedures that refer negative cases to a human
evaluator, using available tools and methods \cite{Hardmeier2016}. Fully
automatic evaluation of a very restricted scope may still be feasible using
test suites designed for specific problems \cite{Bawden2017}.

\section{Acknowledgements}

We would like to thank our annotators, Marie Dubremetz and Miryam de Lhoneux,
for their many hours of painstaking work, and Lesly Miculicich Werlen for
providing the resources necessary to compute the APT scores. The annotation work
was funded by the European Association for Machine Translation. The work carried
out at University of Edinburgh was funded by the ERC H2020 Advanced Fellowship
GA 742137 SEMANTAX and a grant from The University of Edinburgh and Huawei
Technologies. The work carried out at Uppsala University was funded by the
Swedish Research Council under grants 2012-917 and 2017-930.

\bibliography{emnlp2018}

\begin{thebibliography}{18}
\expandafter\ifx\csname natexlab\endcsname\relax\def\natexlab#1{#1}\fi

\bibitem[{Bawden et~al.(2017)Bawden, Sennrich, Birch, and Haddow}]{Bawden2017}
Rachel Bawden, Rico Sennrich, Alexandra Birch, and Barry Haddow. 2017.
\newblock \href {http://arxiv.org/abs/1711.00513} {Evaluating discourse
  phenomena in neural machine translation}.
\newblock \emph{CoRR}, abs/1711.00513.

\bibitem[{Guillou(2015)}]{Guillou2015}
Liane Guillou. 2015.
\newblock Automatic post-editing for the discomt pronoun translation task.
\newblock In \emph{Proceedings of the Second Workshop on Discourse in Machine
  Translation}, pages 65--71, Lisbon, Portugal. Association for Computational
  Linguistics.

\bibitem[{Guillou(2016)}]{GuillouThesis}
Liane Guillou. 2016.
\newblock \emph{Incorporating Pronoun Function into Statistical Machine
  Translation}.
\newblock Ph.D. thesis, Edinburgh University, Department of Informatics.

\bibitem[{Guillou and Hardmeier(2016)}]{Guillou2016}
Liane Guillou and Christian Hardmeier. 2016.
\newblock {PROTEST}: A test suite for evaluating pronouns in machine
  translation.
\newblock In \emph{Proceedings of the Eleventh Language Resources and
  Evaluation Conference}, LREC 2016, pages 636--643, Portoro\v{z}, Slovenia.

\bibitem[{Guillou et~al.(2014)Guillou, Hardmeier, Smith, Tiedemann, and
  Webber}]{ParCor2014}
Liane Guillou, Christian Hardmeier, Aaron Smith, J{\"{o}}rg Tiedemann, and
  Bonnie Webber. 2014.
\newblock {ParCor} 1.0: A parallel pronoun-coreference corpus to support
  statistical {MT}.
\newblock In \emph{Proceedings of the 9th International Conference on Language
  Resources and Evaluation}, LREC 2014, pages 3191--3198, Reykjavik, Iceland.
  European Language Resources Association (ELRA).

\bibitem[{Hardmeier(2014)}]{HardmeierThesis}
Christian Hardmeier. 2014.
\newblock \emph{Discourse in Statistical Machine Translation}.
\newblock Ph.D. thesis, Uppsala University, Department of Linguistics and
  Philology.

\bibitem[{Hardmeier and Federico(2010)}]{Hardmeier2010}
Christian Hardmeier and Marcello Federico. 2010.
\newblock Modelling pronominal anaphora in statistical machine translation.
\newblock In \emph{Proceedings of the 7th International Workshop on Spoken
  Language Translation}, IWSLT 2010, pages 283--289, Paris, France.

\bibitem[{Hardmeier and Guillou(2016)}]{Hardmeier2016}
Christian Hardmeier and Liane Guillou. 2016.
\newblock A graphical pronoun analysis tool for the protest pronoun evaluation
  test suite.
\newblock \emph{Baltic Journal of Modern Computing}, (2):318--330.

\bibitem[{Hardmeier et~al.(2015)Hardmeier, Nakov, Stymne, Tiedemann, Versley,
  and Cettolo}]{Hardmeier2015b}
Christian Hardmeier, Preslav Nakov, Sara Stymne, J{\"o}rg Tiedemann, Yannick
  Versley, and Mauro Cettolo. 2015.
\newblock Pronoun-focused {MT} and cross-lingual pronoun prediction: Findings
  of the 2015 {DiscoMT} shared task on pronoun translation.
\newblock In \emph{Proceedings of the Second Workshop on Discourse in Machine
  Translation}, DiscoMT 2015, pages 1--16, Lisbon, Portugal.

\bibitem[{Hardmeier et~al.(2016)Hardmeier, Tiedemann, Nakov, Stymne, and
  Versely}]{DiscoMT2015TestSet}
Christian Hardmeier, J{\"o}rg Tiedemann, Preslav Nakov, Sara Stymne, and
  Yannick Versely. 2016.
\newblock \href {http://hdl.handle.net/11372/LRT-1611} {{DiscoMT 2015 Shared
  Task on Pronoun Translation}}.
\newblock LINDAT/CLARIN digital library at Institute of Formal and Applied
  Linguistics, Charles University in Prague.

\bibitem[{Jean et~al.(2014)Jean, Cho, Memisevic, and Bengio}]{NYUNMT}
Sébastien Jean, Kyunghyun Cho, Roland Memisevic, and Yoshua Bengio. 2014.
\newblock On using very large target vocabulary for neural machine translation.
\newblock \emph{ArXiv e-prints}, 1412.2007.

\bibitem[{Lo\'{a}iciga and Wehrli(2015)}]{Loaiciga2015}
Sharid Lo\'{a}iciga and Eric Wehrli. 2015.
\newblock Rule-based pronominal anaphora treatment for machine translation.
\newblock In \emph{Proceedings of the Second Workshop on Discourse in Machine
  Translation}, pages 86--93, Lisbon, Portugal. Association for Computational
  Linguistics.

\bibitem[{Loáiciga(2017)}]{LoaicigaThesis}
Sharid Loáiciga. 2017.
\newblock \emph{Pronominal anaphora and verbal tenses in machine translation}.
\newblock Ph.D. thesis, Université de Genève.

\bibitem[{Luong et~al.(2015)Luong, Miculicich~Werlen, and
  Popescu-Belis}]{Luong2015}
Ngoc~Quang Luong, Lesly Miculicich~Werlen, and Andrei Popescu-Belis. 2015.
\newblock Pronoun translation and prediction with or without coreference links.
\newblock In \emph{Proceedings of the Second Workshop on Discourse in Machine
  Translation}, pages 94--100, Lisbon, Portugal. Association for Computational
  Linguistics.

\bibitem[{Mitkov and Barbu(2003)}]{Mitkov:2003}
Ruslan Mitkov and Catalina Barbu. 2003.
\newblock Using bilingual corpora to improve pronoun resolution.
\newblock \emph{Languages in Contrast}, 4(2):201--211.

\bibitem[{Papineni et~al.(2002)Papineni, Roukos, Ward, and Zhu}]{Papineni:2002}
Kishore Papineni, Salim Roukos, Todd Ward, and Wei-Jing Zhu. 2002.
\newblock {BLEU}: A method for automatic evaluation of machine translation.
\newblock In \emph{Proceedings of the 40th Annual Meeting of the Association
  for Computational Linguistics}, pages 311--318, Philadelphia (Pennsylvania,
  USA). ACL.

\bibitem[{Tiedemann(2015)}]{Tiedemann2015}
J\"{o}rg Tiedemann. 2015.
\newblock Baseline models for pronoun prediction and pronoun-aware translation.
\newblock In \emph{Proceedings of the Second Workshop on Discourse in Machine
  Translation}, pages 108--114, Lisbon, Portugal. Association for Computational
  Linguistics.

\bibitem[{Werlen and Popescu-Belis(2017)}]{Werlen2017}
Lesly~Miculicich Werlen and Andrei Popescu-Belis. 2017.
\newblock Validation of an automatic metric for the accuracy of pronoun
  translation ({APT}).
\newblock In \emph{Proceedings of the Third Workshop on Discourse in Machine
  Translation (DiscoMT)}. Association for Computational Linguistics (ACL).

\end{thebibliography}
\bibliographystyle{acl_natbib}

\end{document}